\documentclass[letterpaper]{article} 
\usepackage[arxiv]{aaai24}  
\usepackage{times}  
\usepackage{helvet}  
\usepackage{courier}  
\usepackage[hyphens]{url}  
\usepackage{graphicx} 
\usepackage{natbib}  
\usepackage{caption} 
\usepackage{algorithm}
\usepackage{algorithmic}

\usepackage{newfloat}
\usepackage{listings}
\DeclareCaptionStyle{ruled}{labelfont=normalfont,labelsep=colon,strut=off} 
\floatstyle{ruled}
\newfloat{listing}{tb}{lst}{}
\floatname{listing}{Listing}
\usepackage{color} 
\usepackage{amsmath} 
\usepackage{amsfonts} 
\usepackage{cleveref} 
\usepackage{booktabs} 
\usepackage{multirow} 
\usepackage{subcaption} 
\usepackage{pifont} 
\usepackage[table]{xcolor}
\usepackage{colortbl}
\usepackage{eurosym} 
\usepackage{soul}

\newcommand{\continue}{\textcolor{red}{[...]}}%
\newcommand{\tmp}[1]{\textcolor{red}{#1}}
\newcommand{\at}{\texttt{@}}%
\newcommand{\cmark}{\ding{51}}%
\newcommand{\xmark}{\ding{55}}%
\newcommand{\sub}[1]{\small$_{\textsc{#1}}$}
\newcommand{\sd}[1]{\scriptsize$\pm #1$}
\newcommand{\better}[1]{\scriptsize($+#1\%$)}
\newcommand{\worse}[1]{\scriptsize($-#1\%$)}
\definecolor{shade}{gray}{0.6}
\newcommand{\shade}[1]{\textcolor{shade}{#1}}
\newcommand{\graymidrule}{\arrayrulecolor{lightgray}\midrule\arrayrulecolor{black}}

\definecolor{pastelyellow}{RGB}{253, 253, 150}
\sethlcolor{pastelyellow}

\title{Interpretable Long-Form Legal Question Answering with\\ Retrieval-Augmented Large Language Models}
\author{
    Antoine Louis,
    Gijs van Dijck,
    Gerasimos Spanakis
}
\affiliations{
    Law \& Tech Lab, Maastricht University\\
    \{a.louis, gijs.vandijck, jerry.spanakis\}@maastrichtuniversity.nl
}

\begin{document}
\maketitle

\begin{abstract}
Many individuals are likely to face a legal dispute at some point in their lives, but their lack of understanding of how to navigate these complex issues often renders them vulnerable. The advancement of natural language processing opens new avenues for bridging this legal literacy gap through the development of automated legal aid systems. However, existing legal question answering (LQA) approaches often suffer from a narrow scope, being either confined to specific legal domains or limited to brief, uninformative responses. In this work, we propose an end-to-end methodology designed to generate \textsl{long-form} answers to \textsl{any} statutory law questions, utilizing a ``retrieve-then-read'' pipeline. To support this approach, we introduce and release the Long-form Legal Question Answering (LLeQA) dataset, comprising 1,868 expert-annotated legal questions in the French language, complete with detailed answers rooted in pertinent legal provisions. Our experimental results demonstrate promising performance on automatic evaluation metrics, but a qualitative analysis uncovers areas for refinement. As one of the only comprehensive, expert-annotated long-form LQA dataset, LLeQA has the potential to not only accelerate research towards resolving a significant real-world issue, but also act as a rigorous benchmark for evaluating NLP models in specialized domains. We publicly release our code, data, and models.\footnote{\url{https://github.com/maastrichtlawtech/lleqa}}
\end{abstract}

\section{Introduction \label{sec:introduction}}
Legal disputes are an inevitable part of everyday life, with many individuals finding themselves entangled in issues related to marriage, debts, or employment \citep{farrow2016everyday,ponce2019global}. However,  most people have little to no knowledge about their rights and fundamental legal processes \citep{balmer2010knowledge}. As a result, they either take no action or turn to the internet for advice \citep{denvir2016online}. Unfortunately, the latter often directs users towards commercial websites that prioritize their own marketing efforts over providing thorough, useful legal guidance \citep{hagan2020legal}. While invaluable, expert legal assistance is often prohibitively expensive, which results in a considerable number of vulnerable individuals being left unprotected or exploited due to their inability to afford it. This barrier to accessing legal information fosters a significant imbalance within the legal system, impeding the universal right to equal access to justice for all. The global implications of this issue are significant: an estimated 1.5 billion individuals wrestle with unresolved legal challenges, and 4.5 billion may be excluded from the protective measures that the law provides \citep{garavano2019justice}. In light of these circumstances, there is growing consensus that improved access to legal information could dramatically enhance the outcomes of legal disputes for many people \citep{currie2009legal}.

The rapid progress in natural language processing and the growing availability of digitized legal data present unprecedented opportunities to bridge the gap between people and the law. For instance, legal text summarization \citep{bhattacharya2019comparative,shukla2022legal} holds the potential to simplify complex legal documents for the layperson, while legal judgment prediction (Chalkidis et al. \citeyear{chalkidis2019neural}; Trautmann et al. \citeyear{trautmann2022}) could unveil insightful correlations between an individual's situation and the probable legal outcome. Similarly, legal question answering (LQA) could offer affordable, expert-like assistance to the masses, thereby empowering marginalized parties when utilized for public welfare. However, existing research on LQA tends to exhibit a constrained scope, often concentrating on specialized legal domains, such as tax law (Holzenberger et al. \citeyear{holzenberger2020dataset}) or privacy policies \citep{ravichander2019question}, or limiting the responses to uninformative brief answers like yes/no replies \citep{rabelo2022overview} or few-word spans \citep{duan2019crjc}.

In this paper, we present an end-to-end approach aimed at generating \textsl{long-form} responses to \textsl{any} statutory law questions. Our methodology harnesses the popular ``retrieve-then-read'' pipeline, which first leverages a retriever over a large evidence corpus to fetch a set of relevant legislative articles, and then employs a reader to peruse these articles and formulate a comprehensive, interpretable answer. Our retriever relies on a lightweight bi-encoder model, wich enables fast and effective retrieval. For our reader, we use an instruction-tuned large language model (LLM) that we adapt to our task via two distinct learning strategies: in-context learning, wherein the model learns from instructions and a set of contextually provided examples; and parameter-efficient finetuning, where a small number of extra parameters are optimized on a downstream dataset while the base model's weights are quantized and remain unchanged.

To support training and evaluating such systems, we collect and release the Long-form Legal Question Answering (LLeQA) dataset. LLeQA builds upon BSARD \citep{louis2022statutory}, an information retrieval dataset in French comprising 1,108 legal questions labeled with relevant provisions from a corpus of 22,633 Belgian law articles, and enhance it in two ways. First, we introduce 760 new legal questions (+69\%) and 5,308 additional statutory articles (+23\%). Second, we supplement the data with new types of annotations, including an exhaustive taxonomy for the question, the jurisdictions concerned, the exact paragraph-level references within the relevant articles, and a comprehensive answer written by seasoned legal professionals. Owing to the rich variety of its annotations, LLeQA serves as a multifaceted resource that extends its utility beyond legal question answering and has the potential to catalyze significant progress in various legal tasks, such as legal inquiry classification, legal topic modeling, and legal information retrieval.

Our experimental results show that retrieval-augmented LLMs exhibit commendable performance on automatic evaluation metrics, measuring alignment with target answers. Yet, a deeper qualitative analysis reveals that these syntactically correct responses, despite seemingly covering the intended topics, frequently harbor inaccuracies and erroneous information. This discrepancy underscores the limitations inherent in relying solely on automatic metrics for assessing such systems, and indicates substantial room for improvement both in terms of modeling and evaluation.

In summary, our main contributions are:
\begin{enumerate}
    \item A novel dataset for long-form question answering (LFQA) in the legal domain and French language, comprising 1,868 legal questions, meticulously annotated by legal professionals, with detailed answers and references to relevant legal provisions, drawn from a substantial knowledge corpus containing 27,942 statutory articles.
    \item A comprehensive evaluation of the retrieve-then-read framework in the context of legal LFQA, while emphasizing interpretability and exploring various learning strategies for the reader.
    \item A public release of our code, dataset, and checkpoints to facilitate future research on interpretable legal LFQA.
\end{enumerate}

\begin{table*}[t]
\centering
\resizebox{\textwidth}{!}{%
\begin{tabular}{l|rrr|r|c|c|c|c} 
\toprule
\multirow{2}{*}{\textbf{Dataset}} & \multicolumn{3}{c|}{\textbf{Average \# of words}} & \multirow{2}{*}{\textbf{\# Q-A pairs}} & \multirow{2}{*}{\textbf{Answer type}} & \multirow{2}{*}{\textbf{Domain}} & \multirow{2}{*}{\textbf{Source}} & \multirow{2}{*}{\textbf{Lang.}} \\
 & Question & Evidence & Answer &  &  & &  &  \\ 
\midrule
JEC-QA \citep{zhong2020jec} & 47 & 58 & 15 & 26,365 & Multi-choice & Statutory law & Legal exam & zh \\
SARA (Holzenberger et al. \citeyear{holzenberger2020dataset}) & 46 & 489 & 1 & 376 & Binary, numeric & Tax law & Jurists & en \\
PrivacyQA \citep{ravichander2019question} & 8 & 3,237 & 140 & 1,750 & Multi-span & Privacy policy & Jurists & en \\
CRJC \citep{duan2019crjc} & \textsl{unk.} & \textsl{unk.} & \textsl{unk.} & 51,333 & Binary, span & Case law & Jurists & zh \\
FALQU \citep{mansouri2023falqu} & 144 & - & 244 & 9,880 & Long-form & Statutory law & Online forum & en \\
COLIEE-21 \citep{rabelo2022overview} & 41 & 94 & 1 & 887 & Binary & Civil law & Legal exam & ja, en \\
EQUALS \citep{chen2023equals} & 32 & 252 & 69 & 6,914 & Long-form & Statutory law & Online forum & zh \\ 
\midrule
LLeQA (ours) & 15 & 1,857 & 264 & 1,868 & Long-form & Statutory law & Jurists & fr \\
\bottomrule
\end{tabular}
}
\caption{Comparison of public legal question answering (LQA) datasets. LLeQA has answers an order of magnitude longer and is the only expert-annotated long-form LQA dataset to cover any statutory law subjects.}
\label{tab:lqa_datasets_comparison}
\end{table*}

\section{Related Work \label{sec:related-work}}

\paragraph{Legal question answering.}
Addressing legal questions has long posed intricate challenges within the legal NLP community, stemming from the inherent complexities of legal texts, including specialized terminology, complex structure, and nuanced temporal and logical connections. To stimulate advancement in this field, an array of datasets and benchmarks has emerged. \citet{duan2019crjc} craft a judicial reading comprehension dataset in the Chinese language, aimed at fostering the development of systems capable of mining fine-grained elements from judgment documents. \citet{ravichander2019question} present a corpus of questions about privacy policies of mobile applications with the objective of empowering users to comprehend and selectively investigate privacy matters. Holzenberger et al. (\citeyear{holzenberger2020dataset}) introduce a dataset for statutory reasoning in tax law. \citet{zhong2020jec} present a multi-choice question answering dataset designed to asses professional legal expertise. \citet{rabelo2022overview} hold a competition wherein a task consists in answering ``yes'' or ``no'' given a legal bar exam problem related to Japanese civil law. Lastly, both \citet{mansouri2023falqu} and \citet{chen2023equals} offer a corpus featuring question-answer pairs in English and Chinese, respectively, sourced from online law-oriented forums.

\paragraph{Knowledge-grounded question answering.}
Mainstream approaches to tackling knowledge-intensive QA tasks commonly rely on external knowledge sources to enhance the answer prediction, such as collected documents \citep{voorhees1999trec}, web-pages (Kwok et al. \citeyear{kwok2001scaling}), or structured knowledge bases \citep{berant2013semantic,yu2017improved}. These \textsl{open-book} models (Roberts et al. \citeyear{roberts2020how}) typically index the knowledge corpus before employing a retrieve-then-read pipeline to predict a response based on multiple supporting documents (\citeauthor{chen2017reading} 2017; Lee et al. \citeyear{lee2019latent}; \citeauthor{karpukhin2020dense} 2020). To an extent, this paradigm can be likened to query-based multi-document summarization \citep{tombros1998advantages}, where the objective lies in providing users with a succinct and precise overview of the top-ranked documents related to their queries. Query-driven summarization may adopt different methodologies, manifesting either in an extractive form, where specific portions of evidence text are selected (Otterbacher et al. \citeyear{otterbacher2009biased}, \citeauthor{wang2013sentence} 2013; \citeauthor{litvak2017query} 2017), or in an abstractive form, where the information is synthesized into new expressions (\citeauthor{nema2017diversity} 2017; Baumel et al. \citeyear{baumel2018query}; \citeauthor{ishigaki2020neural} 2020).

\paragraph{Rationale generation.}
To gain insights into model predictions (Lei et al. \citeyear{lei2016rationalizing}), recent advancements have explored the generation of abstractive textual explanations in areas such as commonsense reasoning \citep{rajani2019explain} and natural language inference \citep{kumar2020nile}. Alternatively, \citet{lakhotia2021fidex} proposed the extractive generation of predefined evidence markers instead of decoding raw explanations.  Complementing generation, studies have concentrated on extracting rationales from evidence input segments (Bastings et al. \citeyear{bastings2019interpretable}; \citeauthor{paranjape2020information} 2020; \citeauthor{chalkidis2021paragraph} 2021), as well as analyzing saliency maps to underscore key input tokens instrumental to each prediction (Ribeiro et al. \citeyear{ribeiro2016why}, Murdoch et al. \citeyear{murdoch2018beyond}).

\section{The LLeQA Dataset \label{sec:dataset}}

\subsection{Dataset Construction \label{subsec:dataset_construction}}
In this section, we describe our process to create LLeQA, which involves three main stages. First, we gather and refine annotated legal questions. Then, we build an expansive corpus of supportive statutory articles drawn from Belgian legislation. Finally, we enrich the question annotations by generating paragraph-level references within relevant articles. We elaborate upon each of these steps below.

\paragraph{Collecting question-answer pairs.}
The data construction process starts with collecting high-quality question-answer pairs on a legal matter. Echoing \citet{louis2022statutory}, we partner with Droits Quotidiens, a Belgian non-profit organization  that endeavors to make the law comprehensible and accessible to the most vulnerable. To this end, the organization maintains a rich website featuring thousands of legal questions commonly posed by Belgian citizens. Each question comes with its own individual page, encompassing one or more categorizations, references to relevant legislative statutes, and a detailed answer written in layman's terms by experienced jurists. With their help, we collect approximately 2,550 legal questions. We then filter out questions that are unsuitable for retrieval-based question answering. Specifically, we discard questions whose references are either missing, too vague (e.g., an entire law or book), or from a statute not collected in our knowledge corpus. Additionally, we group duplicate questions found across different subcategories on the website. This yields a final number of 1,868 question-answer pairs, each with legal references.

\paragraph{Mining supporting information.}
Next, we build the knowledge corpus of statutory articles used to provide evidence that a system can draw upon when generating an answer. We start by extracting provisions from all publicly available Belgian codes of law via the official government website, forming an exhaustive foundation of 23,759 articles across 35 legal codes, thereby encapsulating a wide range of legal subjects. To enhance the corpus, we then incorporate additional laws, decrees, and ordinances that are frequently cited as supportive references but absent from the initial collection, adding 4,183 articles from 34 legal acts. This brings the final evidence corpus to 27,942 articles. We proceed by assigning a unique identifier to each article and employ regular expressions to match the plain text legal references linked to the questions with their corresponding article in our corpus. Next, we cleanse the articles of recurrent noisy textual elements, such as nested brackets, superscripts, or footnotes that may be present due to revisions or repeals by new legislation. Besides the article's content, we also collate the complete legislative path leading up to that article, which starts from the statute's name and progresses through the name of the book, act, chapter, section, and subsection where the article resides. This auxiliary information provides valuable contextual insight on the article's subject matter. Lastly, we partition the articles into their constituent paragraphs, which serve as the basic units for rationale extraction.

\paragraph{Generating paragraph-level rationales.}
In our dataset, only 10.4\% of the collected questions come with paragraph-level references, i.e., with mentions to specific article paragraphs as the relevant information to the question. To extend this level of interpretability across all questions in LLeQA, we leverage a large-scale language model to synthetically generate these paragraph-level relevance signals for the remaining samples. This approach serves as an affordable and efficient alternative to hiring costly legal experts for annotation. Utilizing the closed-source gpt-3.5-turbo-0613 model via the OpenAI API, we present the model with all paragraphs from relevant legal articles for each question-answer pair, and instruct it to identify those contributing to the answer. The model responds with a comma-separated list of identifiers corresponding to the deemed relevant paragraphs, or ``None'' if it discerns no contribution. After parsing the model's responses, we incorporate these synthetically generated paragraph-level rationales into the dataset.

\paragraph{Finalizing the dataset.}
We assign the questions with expert-annotated paragraph-level references to the test set and randomly partition the remaining samples into training and development sets, yielding a 10/90/10 split, respectively.

\subsection{Dataset Analysis \label{subsec:dataset_analysis}}
In the following section, we perform an extensive analysis of our dataset to provide insights into its unique characteristics.

\paragraph{Comparative overview.}
\Cref{tab:lqa_datasets_comparison} presents a comparative review of existing LQA datasets, including ours, across several key factors, such as the length of textual elements, number of samples, type of answer, legal domain covered, source of annotations, and language used. LLeQA distinguishes itself by being the only dataset targeting any statutory law subjects, with long-form answers derived from legal experts. Its questions are succinct, intending to mirror a real-world scenario where laypeople may struggle to elaborate on their legal concerns. In contrast, the answers are more detailed than in other datasets, as they often compensate for the lack of provided information by exploring all potential scenarios contingent on an individual's circumstances, such as age, employment situation, or marital status.

\begin{table}[t]
\centering
\resizebox{\columnwidth}{!}{%
\begin{tabular}{lrl} 
\toprule
\textbf{Word} & \textbf{(\%)} & \textbf{Example Question} \\ 
\midrule
Can & 33.9 & Can I continue to work if I am retired? \\
How & 21.2 & How can I end my student lease ? \\
What & 14.8 & What is the role of the guardian of a minor? \\
Must & 8.6 & Must I mention that I am pregnant during a job interview? \\
Who & 3.9 & Who has to pay the funeral expenses? \\
Which & 3.8 & Which expenses are reimbursed in case of work-related accident? \\
When & 1.6 & When do I have to hand in my resignation? \\
Where & 0.5 & Where can I request my criminal record extract? \\
Why & 0.1 & Why do I have to declare the birth of a child to the municipality? \\
OTHER & 11.6 & Do the assets I owned before the marriage become joint assets? \\
\bottomrule
\end{tabular}
}
\caption{LLeQA questions by interrogative word. All examples in the paper are translated from French for illustration.}
\label{tab:lleqa_question_words}
\end{table}

\paragraph{Question diversity.}
In \Cref{tab:lleqa_topic_distribution}, we provide a breakdown of the major question subjects in LLeQA. Housing and healthcare represent the two largest topics, accounting for almost half of all questions together. Family, work, and immigration follow, collectively constituting over a third of the dataset, while money, privacy, and justice questions are less prevalent. We then examine the type of information requested in the questions based on their interrogative words, as shown in \Cref{tab:lleqa_question_words}. ``Can'', ``how'', ``what'', and ``must'' are the most frequently used question words, indicating that people primarily seek information on legal permissions, procedures, definitions, and obligations. Less common are inquiries about identities, specifications, temporal aspects, locational factors, or causal relationships. These distributions reflect the variety of legal concerns that citizens face, thereby providing guidance for user-centric LQA systems.

\paragraph{Question evidence.}
In LLeQA, approximately 80\% of the questions associate with fewer than five articles from the knowledge corpus, with the median number of relevant articles per question being two. These articles have a median length of 84 words, yet 1,515 articles exceed 500 words, with the longest reaching several thousand words. When combining all relevant articles for each question, we find an average evidence length of 1,857 words per question, positioning LLeQA at the upper range among published datasets regarding evidence length. Interestingly, a mere 8\% of the articles within our corpus are referenced as relevant to at least one question within the dataset. Moreover, nearly half of these referenced articles originate from five statutes only, implying the critical role a select few laws play in answering the most frequently posed legal inquiries.

\paragraph{Assessment of annotation quality.}
To assess the quality of the synthetically generated paragraph-level rationales, we evaluate the performance of gpt-3.5-turbo-0613 against the ground truth annotations from the test set. Although far from expert quality, we find that the model demonstrates decent annotation performance, achieving a F1 score of 47.5\%. By comparison, a naive baseline that randomly selects the relevant paragraphs achieves 15.3\% F1, whereas one that always marks the first paragraph as the relevant one scores 27.2\% F1. To ensure that we employ the most suitable model for this annotation task, we experiment with alternative LLMs and present the results in \Cref{tab:annotation_results} of \Cref{app:lleqa_details}. We observe that, as of the time of writing, gpt-3.5-turbo-0613 achieves the best overall performance within a limited cost budget. Despite the apparent margin of error, we believe that these imperfect synthetic annotations may still be beneficial for in-context learning purposes as ground truth labels bear less significance in such settings \citep{min2022rethinking}.

\begin{table}[t]
\centering
\resizebox{0.565\columnwidth}{!}{%
\begin{tabular}{l|ccc|r}
\toprule
\textbf{Topic}  & \textbf{Train} & \textbf{Dev} & \textbf{Test} & \textbf{(\%)} \\ 
\midrule
Housing      & 382            & 54              & 83    & 27.8        \\
Healthcare   & 286            & 40              & 67    & 21.0         \\
Family       & 217            & 22              & 16    & 13.7        \\
Work         & 167            & 26              & 9     & 10.8        \\ 
Immigration  & 156            & 22              & 3     & 9.7       \\
Money        & 120            & 14              & 7     & 7.5       \\
Privacy      & 80             & 14              & 10    & 5.6        \\
Justice      & 64             & 9               & 0     & 3.9       \\
\midrule
Total       & 1472            & 201             & 195   & -        \\ 
\bottomrule
\end{tabular}
}
\caption{Topic distribution of questions in LLeQA.}
\label{tab:lleqa_topic_distribution}
\end{table}

\begin{figure*}[t]
    \centering
    \includegraphics[width=0.94\linewidth]{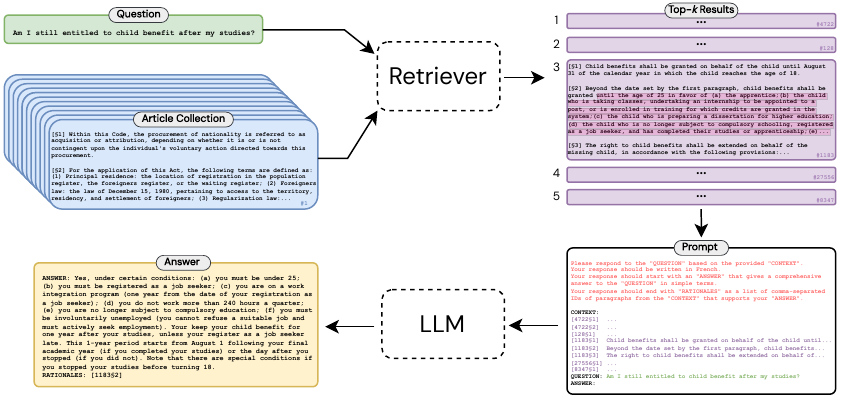}
    \caption{An illustration of the ``retrieve-then-read'' pipeline for interpretable long-form legal question answering.}
    \label{fig:task}
\end{figure*}

\section{Method \label{sec:method}}
In this section, we detail the ``retrieve-then-read'' framework we use for interpretable long-form legal question answering, illustrated in \Cref{fig:task}. First, a \textsl{retriever} selects a small subset of statutory articles, some of which being relevant to the question. Then, a \textsl{generator} conditions its answer on the subset of articles returned by the retriever. We describe these two components in more detail below.

\subsection{Retriever \label{subsec:retriever}}
The role of our retrieval component is to pinpoint all statutory articles relevant to a question and present them at the forefront of the returned results. More formally, the retriever can be expressed as a function $R: (q, \mathcal{C}) \mapsto \mathcal{F}$ that takes as input a question $q$ along with a knowledge corpus of legal provisions $\mathcal{C}=\{p_1, p_2, \cdots, p_N\}$, and returns a notably smaller filtered set $\mathcal{F} \subset \mathcal{C}$ of the supposedly relevant provisions, ordered by decreasing relevance.

We employ the widely adopted bi-encoder architecture \citep{bromley93signature} as the foundation of our retriever, which maps a question $q$ and a legal provision $p$ into dense vector representations and computes a relevance score $s: (q,p) \mapsto \mathbb{R}_+$ between the two by the similarity of their embeddings $\boldsymbol{h}_q, \boldsymbol{h}_p \in \mathbb{R}^d$, i.e.,
\begin{equation}
    s(q,p) = \operatorname{sim}\left(\boldsymbol{h}_q, \boldsymbol{h}_p\right),
\end{equation}
where $\operatorname{sim}: \mathbb{R}^d \times \mathbb{R}^d \rightarrow \mathbb{R}$ is a similarity function, such as the dot product or cosine similarity. These embeddings are typically derived from a pooling operation on the output representations of a pretrained autoencoding language model, such as BERT \citep{devlin2019bert}, so that
\begin{equation}
\begin{split}
    \boldsymbol{h}_{q} &= \operatorname{pool}\left(E(q;\theta_1)\right), \text{and}\\
    \boldsymbol{h}_p &= \operatorname{pool}\left(E(p;\theta_2)\right),
\end{split}
\end{equation}
where the model $E(\cdot; \theta_i): \mathcal{W}^{n} \rightarrow \mathbb{R}^{n \times d}$ with weights $\theta_i$ transforms an input text sequence of $n$ terms from vocabulary $\mathcal{W}$ to $d$-dimensional real-valued word vectors. The pooling function $\operatorname{pool}: \mathbb{R}^{n \times d} \rightarrow \mathbb{R}^{d}$ implements a mean or max operation on the output word embeddings to distill a global representation for the text passage. Beyond the conventional \textsl{two-tower} bi-encoder \citep{karpukhin2020dense,yang2020multilingual}, which employs two independent encoder models to map queries and articles separately into distinct embedding spaces, the \textsl{siamese} variant \citep{reimers2019sentence,xiong2021approximate} uses a unique encoder, i.e., $\theta_1 = \theta_2$, to encode the question and article in a shared dense vector space. We use the latter variant in this work.

The training objective for our dense retriever is to learn a high-quality low-dimensional embedding space for questions and legal provisions such that relevant question-provision pairs appear closer to each other than irrelevant ones. Assume our training data $\mathcal{D}=\{\langle q_{i}, p_{i}^{+}\rangle\}_{i=1}^{N}$ contains $N$ instances, each comprising a question $q_{i}$ linked to a relevant provision $p_{i}^{+}$. For each question $q_i$, we sample a set of irrelevant provisions $\mathcal{P}_{i}^{-}$, thereby constituting a training set $\mathcal{T}=\{\langle q{i}, p_{i}^{+}, \mathcal{P}_{i}^{-}\rangle\}_{i=1}^{N}$. Subsequently, we use the instances in $\mathcal{T}$ to contrastively optimize the negative log-likelihood of the relevant provision against the non-relevant ones, i.e.,
\begin{equation}
\label{eq:loss}
    \mathcal{L}_\theta\left(q_{i}, p_{i}^{+}, \mathcal{P}_{i}^{-}\right) = -\log \frac{e^{s(q_i,p_{i}^{+})/\tau}}{\sum_{p \in \mathcal{P}_{i}^{-} \cup \{p_{i}^{+}\}} e^{s(q_{i}, p)/\tau}},
\end{equation}
where $\tau > 0$ is a temperature parameter that we set to 0.01. To select irrelevant provisions, we employ two distinct sampling strategies: \textsl{random sampling} using in-batch negatives \citep{chen2017sampling, henderson2017efficient}, which considers provisions paired with the other questions within the same mini-batch; and \textsl{hard negative sampling} using BM25 \citep{robertson1994okapi}, which includes the top provisions returned by BM25 that bear no relevance to the question.

\subsection{Generator \label{subsec:generator}}
Our generator aims at formulating a comprehensive answer to a short legal question, leaning on corroborative data. Formally, the task can be cast as a conditional text generation problem, where the model requires conditioning its response on a context string that incorporates the question and several supporting statutory articles. We turn to autoregressive large language models (LLMs) based on the Transformer's decoder block \citep{vaswani2017attention} as the backbone architecture for our generator. We then delve into two learning scenarios, specifically in-context learning \citep{radford2019language,brown2020language} and parameter-efficient finetuning \citep{lester2021power,liu2022prompt}.

\paragraph{In-context learning.}
To assess the innate performance of our generator without additional training, we start by examining three prevalent in-context learning strategies, namely \textsl{zero-shot}, \textsl{one-shot}, and \textsl{few-shot} learning. Under the zero-shot learning paradigm, the generator receives a natural language instruction and seeks to answer the question directly. The context $c$ provided to the model is formulated as
\begin{equation}
    c = \left[d, \mathcal{P}_{t}^{+}, q_t\right],
\end{equation}
where $d$ stands for the task description, $q_t$ is the test question, and $\mathcal{P}_{t}^{+}$ represents the top-$k$ most pertinent legal provisions to $q_t$ as identified by the retriever ($k$ is set to 5). In one-shot learning, the generator benefits from an additional demonstration to guide its understanding of the task, whereas in a few-shot setting, the model accommodates as many demonstrations as can be included within its context window. Under these scenarios, the context string becomes
\begin{equation}
    c = \left[d, \left[\mathcal{P}_{j}^{+}, q_j, a_j\right]_{j=1}^{n},  \mathcal{P}_{t}^{+}, q_t\right],
\end{equation}
where $a_j$ is the gold long-form answer to the training question $q_j$, and $n$ denotes the number of demonstrations. We dynamically select demonstrations in the training pool through a similarity-based retrieval method based on the question. 


\paragraph{Parameter-efficient finetuning.}
Due to the gigantic size of contemporary LLMs, performing full finetuning of all model parameters would be prohibitively expensive. Instead, we employ parameter-efficient finetuning to train our generator, which significantly trims both the training duration and computational cost. Specifically, we apply the QLoRA technique \citep{dettmers2023qlora}, which undertakes a preliminary quantization of the pretrained model to 4-bit before freezing all its parameters for training. A small set of learnable low-rank adapter weights \citep{hu2022lora} are then injected into each linear layer of the Transformer and tuned by backpropagating gradients through the quantized weights. Formally, given a training sample $(q_i, \mathcal{P}_{i}^{+}, a_i)$ where $a_i = (y_1, \cdots, y_T)$ represents the target output sequence, we optimize the adapter's parameters $\phi$ using a standard language modeling objective that maximizes the negative log likelihood of generating the target answer $a_i$ conditioned on the input context, encompassing the source question $q_i$ and the set of relevant legal provisions $\mathcal{P}_{i}^{+}$, i.e.,
\begin{equation}
\begin{split}
    \mathcal{L}_\phi\left(q_i, \mathcal{P}_{i}^{+}, a_i\right) 
    &= -\log p_\phi\left(a_i \mid \mathcal{P}_{i}^{+}, q_i\right) \\
    &=-\log \prod_{t=1}^{T} p_\phi\left(y_t \mid \mathcal{P}_{i}^{+}, q_i, y_{<t}\right).
\end{split}
\end{equation}

\paragraph{Context window extension.}
LLMs typically come with a predefined context window limit, beyond which perplexity steeply rises due to the weak extrapolation properties of positional encoding. This limitation poses significant challenges for applications requiring the processing of extensive inputs, like ours. Recent efforts have aimed to extend the context window sizes of pretrained LLMs employing rotary position embedding \citep[RoPE]{su2021roformer}, such as LLaMA \citep{touvron2023llama}, by interpolating positional encoding \citep{chen2023extending}. Guided by promising findings from the open-source community, we perform dynamic NTK-aware scaling,\footnote{\url{https://reddit.com/r/LocalLLaMA/comments/14mrgpr/}} which retains the exact position values within the original context window and progressively down-scales the input position indices using a nonlinear interpolation from neural tangent kernel theory (Jacot et al. \citeyear{jacot2018neural}). Preliminary results suggest this approach substantially mitigates perplexity degradation for sequences exceeding the maximum window size, without necessitating additional finetuning.

\paragraph{Rationales extraction.}
Given the serious implications of flawed legal guidance, ensuring interpretability in generated answers is crucial. This enables users to cross-verify responses through reliable sources while understanding the underlying reasoning, thereby enhancing the trustworthiness of LQA systems. To this end, we impose additional constraints on the model to furnish proper justification for its answers. While prior work on rationale generation has predominantly focused on creating free-form natural language explanations  \citep{latcinnik2020explaining,narang2020wt5}, abstractive models have shown a propensity for fabricating convincing yet misleading justifications, inadvertently supporting inaccurate predictions (\citeauthor{camburu2020make} \citeyear{camburu2020make}; Wiegreffe et al. \citeyear{wiegreffe2021measuring}). Besides, adapting this strategy for applications involving multiple extensive evidence documents proves challenging. Consequently, we adopt an extractive rationale generation strategy \citep{lakhotia2021fidex}, prompting the model to generate evidence paragraph markers rather than raw explanations. This technique ensures the production of unaltered rationales that are easily interpretable.

\section{Experiments \label{sec:experiments}}

\subsection{Experimental Setup \label{subsec:experimental_setup}}

\paragraph{Models.}
Our dense retrieval model leverages CamemBERT \citep{martin2019camembert}, a leading autoencoding model for the French language. To furnish a well-rounded comparative analysis, we incorporate two robust retrieval baselines: BM25 \citep{robertson1994okapi}, a widely utilized bag-of-words retrieval function; and mE5 \citep{wang2022text}, currently the top-performing multilingual dense model on the MTEB benchmark \citep{muennighoff2023mteb}.
With regard to our generator, we experiment with several instruction-tuned open-source models derived from LLaMA \citep{touvron2023llama} to harness the benefits of dynamic NTK-aware scaled RoPE. Specifically, we consider four models that are notably high-ranking on the MT-Bench leaderboard \citep{zheng2023judging}: Vicuna-1.3 \citep{chiang2023vicuna}, WizardLM-1.0 \citep{xu2023wizardlm}, TÜLU \citep{wang2023how}, and Guanaco \citep{dettmers2023qlora}. Due to limited computational resources, we restrict our study to their 7B variant and curtail the extended context window size to 8192 tokens for in-context learning and 4096 tokens for finetuning.

\paragraph{Implementation.}
We start by fully finetuning CamemBERT on LLeQA with a batch size of 32 and a maximum sequence length of 384 tokens for 20 epochs (i.e., 5.8k steps), using AdamW \citep{loshchilov2017decoupled} with $\beta_1=0.9$, weight decay of 0.01, and learning rate warm up along the first 60 steps to a maximum value of 2e-5, after which linear decay is applied. We use 16-bit automatic mixed precision to accelerate training, which takes about 1.7 hours. We then optimize our baseline LLMs through 4-bit QLoRA finetuning with an effective batch size of 8 for 10 epochs (i.e., 1.1K steps) using paged AdamW optimizer with default momentum parameters and constant learning rate schedule of 2e-4. We employ NormalFloat4 with double quantization for the base models and add LoRA adapters on all linear layers by setting $r=16$, $\alpha=32$ while utilizing float16 as computation datatype. Training takes around 7.5 hours to complete. We generate from the LLMs using nucleus sampling \citep{holtzman2020curious} with $p=0.95$ and a temperature of $0.1$.

\paragraph{Hardware \& Libraries.}
Computations are performed on a single 32GB NVIDIA V100 GPU hosted on a server with a dual 20-core Intel Xeon E5-2698 v4 CPU and 512GB of RAM, operating under Ubuntu 16.04. Our code relies on SBERT \citep{reimers2019sentence}, Transformers \citep{wolf2020transformers}, PEFT \citep{mangrulkar2022peft}, FastChat \citep{zheng2023judging}, and Wandb \citep{biewald2020wandb}.

\subsection{Automatic Evaluation \label{subsec:automatic_evaluation}}
\paragraph{Evaluation metrics.}
To evaluate our framework's effectiveness, we assess three core aspects: \textsl{retrieval performance}, \textsl{generation quality}, and \textsl{rationales accuracy}. Firstly, it is essential that the retriever returns as many pertinent provisions as possible within the first top-$k$ results, given the generator's limited context window size. This requirement implies a primary interest in recall at small cutoffs. Additionally, we report the mean reciprocal rank, as it offers valuable insights into the position of the first relevant result. Gauging the quality of long-form answers presents more intricate challenges. Automatic metrics, such as ROUGE \citep{lin2004rouge}, proved to be inadequate due to the intrinsically open-ended nature of long-form responses, which allows for an array of possible formulations that retain semantic similarity \citep{wang2022modeling,xu2023critical}. Besides, these lexical overlap metrics fail to assess essential aspects such as factual correctness and query relevance. Ultimately, a thorough assessment of such systems requires human evaluation, although the latter introduces its own set of challenges (Krishna et al. \citeyear{krishna2021hurdle}), in addition to being expensive and noisy. Owing to resource constraints in our study, we opt for an automated evaluation metrics, fully cognizant of its limitations, and earmark human evaluation for future work. In particular, we report METEOR \citep{banerjee2005meteor}, which demonstrated superior correlation with human judgment in the context of long text generation \citep{sharma2017relevance, chen2022mtg}. Lastly, we evaluate the model's capacity for rationale extraction by measuring the F1 score over the set of predicted paragraph markers as compared to the ground truth markers.

\begin{table}[t]
\centering
\resizebox{\columnwidth}{!}{%
\begin{tabular}{ll|c|ccc}
\toprule
&\textbf{Model}      & \textbf{\#Param} & \textbf{R@5} & \textbf{R@10} & \textbf{MRR@10} \\ 
\midrule
\multicolumn{6}{l}{\textbf{Baselines}} \\
\shade{1} & BM25                      &  -      & 17.4 & 22.8 & 22.0 \\
\shade{2} & mE5\sub{base}             &  278.0M & 15.4 & 21.7 & 25.8 \\
\shade{3} & mE5\sub{large}            &  559.9M & 16.5 & 26.7 & 28.3 \\
\midrule
\multicolumn{6}{l}{\textbf{Ours}}  \\
\shade{4} & CamemBERT\sub{base}       &  110.6M & \textbf{48.6} \sd{1.8} & \textbf{60.6} \sd{2.0} & \textbf{60.0} \sd{1.4} \\
\bottomrule
\end{tabular}
}
\caption{Retrieval scores of our dense retriever benchmarked against other strong retrieval baselines on LLeQA dev set, averaged across five runs with different random seeds.}
\label{tab:retrieval_results}
\end{table}

\paragraph{Results.}
As depicted in \Cref{tab:retrieval_results}, our dense retriever, fine-tuned on a mere 1.5k in-domain examples, significantly outperforms robust retrieval baselines, underlining the essential role of domain adaptation in enhancing performance. However, the results leave substantial room for improvement; on average, less than half of the relevant articles appear within the top five returned results. This shortfall represents a major bottleneck for our generator, which faces the challenge of answering questions based on partially irrelevant provisions. The impact of this limitation is palpable in the poor performance of rationale extraction, which approaches near-zero effectiveness for most LLMs, though part of this deficiency can be attributed to the models' tendency to hallucinate. In terms of answer quality, WizardLM and Vicuna display a high degree of overlap with the ground truth responses, indicating accurate engagement with the subject matter. Providing a demonstration appears to improve generation quality as compared to a zero-shot setup, except for Guanaco which shows strong zero-shot results. However, performance does not seem to vary significantly when more demonstrations are provided. Finally, the results suggest that finetuning on our task-specific dataset consistently enhances model performance across all generators.

\begin{table}[t]
\centering
\resizebox{\columnwidth}{!}{%
\begin{tabular}{ll|rrrr|rrrr}
\toprule
&\multicolumn{1}{c}{\textbf{Task} $(\rightarrow)$}  & \multicolumn{4}{c}{\textbf{Generation (METEOR)}}       & \multicolumn{4}{c}{\textbf{Rationale extraction (F1)}} \\ \cmidrule(r){2-2} \cmidrule(lr){3-6} \cmidrule(lr){7-10}
&\multicolumn{1}{c}{\textbf{Model} $(\downarrow)$}  & \textbf{0-S} & \textbf{1-S} & \textbf{F-S} & \multicolumn{1}{c}{\textbf{FT}} & \textbf{0-S} & \textbf{1-S} & \textbf{F-S} & \textbf{FT}\\ 
\midrule
\shade{1} & Vicuna      & 11.6 & 16.2 & 15.3 & 19.7 & 0.4 & 0.6 & 0.2 & 0.0 \\
\shade{2} & WizardLM    & 12.3 & 15.5 & 16.6 & 20.4 & 0.0 & 0.0 & 0.0 & 2.0 \\
\shade{3} & TÜLU        &  2.9 &  4.6 &  8.5 & 12.7 & 0.1 & 0.0 & 0.0 & 3.5 \\
\shade{4} & Guanaco     & 11.2 & 11.2 & 11.3 & 20.1 & 1.3 & 0.4 & 0.0 & 0.0 \\
\bottomrule
\end{tabular}
}
\caption{Performances of our baseline LLMs on LLeQA test set. Models are evaluated in four learning setups: \textsl{zero-shot} (0-S), \textsl{one-shot} (1-S), \textsl{few-shot} (F-S), and \textsl{finetuned} (FT).}
\label{tab:generation_results}
\end{table}

\subsection{Qualitative Analysis}
To discern the strengths and shortcomings of our generators, we conduct a detailed manual analysis of 10 randomly selected samples from the test set. We find that TÜLU exhibits a propensity for producing concise answers, a tendency likely due to its extensive finetuning on instruction datasets averaging a relatively short completion length of 98.7 words, which also explains its low METEOR score as LLeQA answers are markedly longer. We further observe that Guanaco and WizardLM are prone to repetitiveness in their responses, occasionally echoing identical phrases. While presence and frequency penalties could mitigate this issue, they may prove ineffective in instances addressing specialized topics where the same terminology is intrinsically repeated. Regarding response quality, WizardLM and Vicuna stand out, far exceeding Guanaco, which tends to produce nonsensical or linguistically convoluted sentences. In contrast, WizardLM and Vicuna's outputs are articulated in impeccable French, displaying a persuasive flair that could potentially mislead an unsuspecting reader. Nevertheless, a deeper probe unveils striking hallucinations \citep{ji2023survey}. Despite seemingly addressing the question, many facts, dates, sources, and conditions appear to be fabricated, as if the models leveraged the provided context less for factual accuracy and more as a foundation upon which to weave a convincing fictitious answer.

\section{Conclusion \label{sec:conclusion}}
In this work, we introduce LLeQA, an expert-annotated dataset tailored to facilitate the development of models aimed at generating comprehensive answers to legal questions, while supplying interpretable justifications.  We experiment with the ``retrieve-then-read'' pipeline on LLeQA and explore various state-of-the-art large language models as readers, that we adapt to the task using several learning strategies. We find that this framework tends to produce syntactically correct answers pertinent to the question's subject matter but occasionally fabricate facts. Overall, we believe LLeQA can serve as a robust foundation for advancements in interpretable, long-form legal question answering, thereby contributing to the democratization of legal access.

\paragraph{Limitations and future work.}
Despite our efforts to make LQA systems more factually grounded in supporting legal provisions, the framework we propose remains vulnerable to hallucinations in both the constructed answers and associated rationales. Additionally, consistent with prior studies \citep{krishna2021hurdle,xu2023critical}, we observe that conventional automatic metrics may not accurately mirror answer quality, leading to potential misinterpretations. These challenges present great avenues for future work.

\paragraph{Ethical considerations.}
The premature deployment of LQA systems poses a tangible risk to laypersons, who may uncritically rely on the furnished guidance and inadvertently exacerbate their circumstances. To ensure the responsible development of legal aid technologies, we are committed to limiting the use of our dataset strictly to research purposes.

\section*{Acknowledgments}
This research is partially supported by the Sector Plan Digital Legal Studies of the Dutch Ministry of Education, Culture, and Science. In addition, this research was made possible, in part, using the Data Science Research Infrastructure (DSRI) hosted at Maastricht University.

\bibliography{refs.bib}

\appendix

\section{Details of LLeQA \label{app:lleqa_details}}
\paragraph{Documentation.}
To supplement the dataset description delineated in \Cref{sec:dataset}, we incorporate a \textsl{dataset nutrition label} for LLeQA, as presented in \Cref{tab:data_nutrition_labels} \citep{holland2018dataset}. This label serves as a diagnostic tool, offering a concise overview of crucial dataset characteristics through a standardized format. Its purpose is to facilitate a rapid assessment of the dataset's suitability and alignment with the specific requirements of AI model development.

\paragraph{Dataset examples.}
\Cref{tab:lleqa_examples} illustrates some samples from the LLeQA dataset. A system is tasked with generating comprehensive, long-form answers to short legal questions, utilizing the supportive statutory articles provided. Furthermore, the system is required to offer an appropriate rationale for its responses by pinpointing the relevant paragraphs that underpin its conclusions.

\begin{figure}[t]
    \centering
    \includegraphics[width=1\columnwidth]{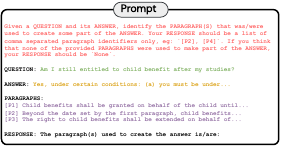}
    \caption{Template for paragraph-level rationale extraction.}
    \label{fig:annotation_prompt}
\end{figure}

\paragraph{Automatic annotation.}
\Cref{fig:annotation_prompt} illustrates the prompt template we use to synthetically generate paragraph-level rationales. As shown in \Cref{tab:annotation_results}, we experiment with various closed-source LLMs available via API, including Command by Cohere, Claude by Anthropic AI, Jurassic by AI21 Labs, and both ChatGPT and GPT-4 by OpenAI. Additionally, we explore the 30B variants of the open-source models delineated in \Cref{subsec:experimental_setup}. Among the evaluated models, GPT-4 emerges as the superior performer, achieving an impressive 65.5\% F1 score. Nevertheless, its financial cost poses a challenge: the annotation of our training set with GPT-4 would approximate \euro{}750, a figure ten times higher than the cost incurred using ChatGPT. Constrained by our budget, we opt for the latter model and defer to future work the development of an enhanced version of these annotations.

\section{Evaluation Metrics \label{app:evaluation_metrics}}
In the following, we formally describe the evaluation metrics we use in our experiments, as motivated in \Cref{subsec:automatic_evaluation}.

\paragraph{Retrieval performance.}
Let $\operatorname{rel}_{q}(p) \in \{0,1\}$ be the binary relevance label of a legal provision $p$ for a specific question $q$, and $\langle i, p\rangle \in \mathcal{F}_q$ a result tuple from the filter set $\mathcal{F}_q \subset \mathcal{C}$ of ranked provisions retrieved for question $q$, with $p$ appearing at rank $i$. The recall for query $q$ is defined as the proportion of relevant provisions within $\mathcal{F}_q$ to the total relevant provisions for $q$ in the entire knowledge corpus $\mathcal{C}$:
\begin{equation*}
    \operatorname{R}_{q}=\frac{\sum_{\langle i, p\rangle \in \mathcal{F}_q} \operatorname{rel}_{q}(p)}{\sum_{p \in \mathcal{C}} \operatorname{rel}_{q}(p)}.
\end{equation*}
In contrast, the reciprocal rank quantifies the inverse of the rank of the first retrieved relevant provision, i.e.,
\begin{equation*}
    \operatorname{RR}_{q}=\max _{\langle i, p\rangle \in \mathcal{F}_q} \frac{\operatorname{rel}_{q}(p)}{i}.
\end{equation*}
These metrics are typically evaluated on a filter set with a specific cutoff $k=\left|\mathcal{F}_q\right|$, much smaller than the corpus size $\left|\mathcal{C}\right|$. We report the \textsl{macro-averaged} recall at cutoff $k$ (R@$k$), and the \textsl{mean} reciprocal rank (MRR), both of which are calculated as average values over a collection of $N$ questions.

\begin{table}[th!]
\centering
\resizebox{\columnwidth}{!}{%
\begin{tabular}{l} 
\toprule
\textbf{Question}: Is it legal to sign a life lease?
\\[0.1cm]
\textbf{Taxonomy}: Housing $\rightarrow$ Rental $\rightarrow$ Residence lease.
\\[0.1cm]
\textbf{Jurisdictions}: Brussels-Capital, Wallonia.
\\[0.1cm]
\graymidrule
\textbf{Answer}: Yes, you can enter into a lifetime lease. Here, we are\\ referring to your life, that of the tenant. The lease automatically\\ ends at the time of your death. You can also enter into a lifetime\\ lease with several other tenants. It ends upon the death of the last\\ surviving tenant. A lifetime lease is rarely used. A lease based on\\ the life of the owner is not possible.
\\[0.1cm]
\textbf{Rationales}: 2517§2.
\\[0.1cm]
\graymidrule
\textbf{Article \#2517}: \\
$[$§1$]$ Notwithstanding Art. 237, a lease can also be entered into in\\ writing, for a duration longer than nine years. This lease ends at\\ the expiry of the agreed term, provided a notice is given by either\\ party at least six months before the deadline. ...\\
$[$§2$]$ Notwithstanding Art. 237, a written lease may be concluded\\ for the life of the lessee. The lease ends automatically upon the\\ death of the lessee. This lease is not governed by ... \\
\midrule
\textbf{Question}: Can I skip my last two months' rent to make sure I get\\ my rental deposit back?
\\[0.1cm]
\textbf{Taxonomy}: Housing $\rightarrow$ Rental $\rightarrow$ Residence lease $\rightarrow$ Rental deposit.
\\[0.1cm]
\textbf{Jurisdictions}: Brussels-Capital, Wallonia.
\\[0.1cm]
\graymidrule
\textbf{Answer}: No. These are two different things. You can't decide by\\ yourself not to pay the last two months' rent because you've paid\\ a rental deposit, nor can you break the lease two months before the\\
end of the term on that basis alone. However, if your landlord agr-\\ees, there's no problem. In that case, put your agreement in writing.
\\[0.1cm]
\textbf{Rationales}:  25788§3.
\\[0.1cm]
\graymidrule
\textbf{Article \#25788}: \\
$[$§1$]$ If, regardless of the securities provided in article 20 of this\\ decree, the tenant provides to ensure the respect of their obliga-\\tions, one of the forms of guarantees provided in the following\\ paragraph, it cannot exceed an amount equivalent to 2 or 3 mo-\\nths of rent, depending on the form of the rental guarantee. ...\\
$[$§2$]$ When the landlord is in possession of the guarantee and re-\\frains from placing it in the manner provided in § 1, 3°, they are\\ obligated to pay the tenant interest at the average financial mar-\\ket rate on the amount of the guarantee, from the time of its han-\\dover. These interests are capitalized. ... \\
$[$§3$]$ There can be no disposal of the bank account, both in prin-\\cipal and interest, or of the bank guarantee, nor of the account on\\ which the guarantee was replenished, except for the benefit of one\\ or the other party, by providing either a written agreement, establ-\\ished at the earliest at the end of the lease contract, or a copy of a\\ court decision. ...\\
\bottomrule
\end{tabular}
}
\caption{Examples from the LLeQA dataset.}
\label{tab:lleqa_examples}
\end{table}

\begin{table*}[!ht]
\centering
\small
\begin{subtable}[t]{.48\linewidth}
\centering
\caption*{}
    \begin{tabular}{|llr|} 
    \hline
    \multicolumn{3}{|l|}{\textbf{\Large Data Facts}} \\
    &  \multicolumn{2}{l|}{Long-form Legal Question Answering (LLeQA) Dataset} \\
    \multicolumn{3}{|l|}{} \\
    \hline
    \multicolumn{3}{l}{} \\ 
    \hline
    \multicolumn{3}{|l|}{\textbf{\large Metadata}} \\ 
    \hline
    & \textbf{Filenames} & articles$^\star$, questions\_\{train$^\ast$, dev$^{\dagger}$, test$^{\ddagger}$\} \\ 
    \hline
    & \textbf{Size} & 27942$^\star$, 1472$^\ast$, 201$^{\dagger}$, 195$^{\ddagger}$ \\ 
    \hline
    & \textbf{Variables} & 14$^\star$, 7$^{\ast\dagger\ddagger}$ \\ 
    \hline
    & \textbf{Domain} & law \\ 
    \hline
    & \textbf{Language} & French (fr-BE) \\ 
    \hline
    & \textbf{Uses} & question answering, information retrieval \\ 
    \hline
    & \textbf{Format} & json \\ 
    \hline
    & \textbf{Type} & nested \\ 
    \hline
    & \textbf{License} & CC BY-NC-SA 4.0 \\ 
    \hline
    & \textbf{Collected} & March 2023 \\ 
    \hline
    & \textbf{Released} & August 2023 \\ 
    \hline
    & \textbf{Description} & This dataset is a collection of short legal\\
    &                      & question and statutory articles from the\\ 
    &                      & Belgian legislation. Each question is\\ 
    &                      & annotated with a comprehensive answer\\
    &                      & and references to relevant articles from\\
    &                      & the legislation. The annotations were made\\
    &                      & by a team of experienced Belgian jurists. \\
    \hline
    \multicolumn{3}{l}{} \\ 
    \hline
    \multicolumn{3}{|l|}{\textbf{\large Provenance}} \\ 
    \hline
    & \textbf{Source \#1} &  \\
    & \quad Name & Belgian consolidated legislation \\
    & \quad Url & \url{https://www.ejustice.just.fgov.be/loi/loi.htm} \\
    & \quad Email & justel@just.fgov.be \\ 
    \hline
    & \textbf{Source \#2} &  \\
    & \quad Name & Droits Quotidiens ASBL \\
    & \quad Url & \url{https://www.droitsquotidiens.be} \\
    & \quad Email & info@droitsquotidiens.be \\ 
    \hline
    & \textbf{Author} &  \\
    & \quad Name & Anonymous \\
    & \quad Url & https://anonymous.com \\
    & \quad Email & anonymous@mail.com \\
    \hline
    \end{tabular}
\end{subtable}%
\begin{subtable}[t]{.53\linewidth}
\centering
\caption*{}
    \begin{tabular}{|lr|}
        \hline
        \textbf{\large Variables} & \\ 
        \hline
        \quad \textbf{articles} & \\
        \qquad id & A unique integer for the article. \\
        \qquad article & The full article content. \\
        \qquad description & The concatenated section headings. \\
        \qquad authority & The legal authority for the article. \\
        \qquad reference & The full name of the article. \\
        \qquad statute & The name of the parent statute. \\
        \qquad article\_no & The article number in that statute. \\
        \qquad book & The name of the parent book, if any. \\
        \qquad part & The name of the parent part, if any. \\
        \qquad act & The name of the parent act, if any. \\
        \qquad chapter & The name of the parent chapter, if any. \\
        \qquad section & The name of the parent section, if any. \\
        \qquad subsection & The name of the parent subsection, if any. \\
        \qquad paragraphs & The distinct paragraphs from the article. \\
        \hline
        \quad \textbf{questions} & \\
        \qquad id & A unique integer for the question. \\
        \qquad question & The question text. \\
        \qquad answer & A detailed answer to the question. \\
        \qquad regions & The jurisdictions in Belgium concerned. \\
        \qquad topics & A taxonomy of related topics. \\
        \qquad article\_ids & The IDs of the relevant articles.\\
        \qquad paragraph\_ids & The relevant paragraph IDs in these articles.\\
        \hline
    \end{tabular}
\end{subtable}
\caption{Dataset Nutrition Label \citep{holland2018dataset} of LLeQA.}
\label{tab:data_nutrition_labels}
\end{table*}

\paragraph{Generation quality.}
The METEOR metric \citep{banerjee2005meteor} serves as a sophisticated word-based evaluation tool to assess system generation quality by comparing the overlap between generated words and a reference text. Distinct from a mere exact word matching approach, METEOR employs an incremental word alignment method that initially identifies exact word-to-word mappings, then extends to stemmed-word matches, and ultimately considers synonym and paraphrase correspondences. Once an alignment is constructed between the generated and reference texts, unigram precision (P) and recall (R) are calculated. The precision is the proportion of matched words in the generation to the total words in that generation, while recall is the ratio of matched words in the generation to the total words in the reference. These are then combined via an harmonic mean, placing most of the weight on recall, as per the formula:
\begin{equation*}
   \operatorname{F-score} = \frac{\operatorname{10PR}}{\operatorname{9P}+\operatorname{R}}.
\end{equation*}
Since the previous measure emerges solely from unigram matches, METEOR seeks to incorporate longer matches by assessing a penalty for a particular alignment. Hence, matched ``chunks'' (i.e., contiguous unigrams, or n-grams) are identified in the generation relative to the reference. Longer n-gram matches result in fewer chunks, with a single chunk indicating a complete match between the generation and reference. In the absence of bigram or longer matches, the number of chunks equals the number of unigrams. The penalty is computed using the minimum possible number of chunks in a generation, represented as
\begin{equation*}
    \lambda = 0.5 \left( u/n \right)^3,
\end{equation*}
where $u$ denotes the number of matched unigrams, and $n$ is the count of matched n-grams. The final score is derived as
\begin{equation*}
    \text{METEOR} = (1 - \lambda)\cdot\operatorname{F-score},
\end{equation*}
which has the effect of diminishing the F-score by up to 0.5 if there are no n-gram matches.

\paragraph{Rationale accuracy.}
The F$_1$ score provides a measure of the accuracy of predicted paragraph markers in comparison to the ground truth markers. Formally, it is defined as
\begin{equation*}
\operatorname{F}_1 = \frac{2\operatorname{PR}}{\operatorname{P}+\operatorname{R}},
\end{equation*}
where precision (P) is the ratio of the count of correctly predicted markers to the total number of predicted markers, and recall (R) represents the ratio of the count of correctly predicted markers to the total number of ground truth markers.

\begin{table*}[t]
\centering
\resizebox{\textwidth}{!}{%
\begin{tabular}{l} 
\toprule
\textbf{Question}: Comment envoyer mon préavis au propriétaire en Wallonie ?\\
\midrule
\textbf{Gold answer}: Vous devez donner votre préavis soit : par courrier recommandé ; via un huissier de justice ; directement entre les mains du propriétaire.\\Il doit signer le double et vous le remettre. La date de réception doit être indiquée. Mais vous pouvez toujours vous mettre d'accord avec le propriétaire\\ sur d'autres modalités pour envoyer le préavis. Il est préférable de prévoir ces autres modalités dans le contrat de bail. Attention, si vous avez signé le\\ bail avec d'autres personnes, toutes ces personnes doivent signer le préavis pour qu'il soit valable. Une exception, si votre bail est un bail de colocation,\\ vous pouvez envoyer seul votre préavis à certaines conditions. Si vous êtes mariés ou cohabitants légaux, votre époux ou cohabitant légal doit aussi\\ signer le préavis, même s'il n'avait pas signé le bail.\\
\graymidrule
\textbf{Zero-shot}: Pour envoyer un préavis au propriétaire en Wallonie, vous devez suivre les procédures légales en matière d'envoi de documents judiciaires.\\Vous devez délivrer le préavis au moins six mois avant l'échéance, et le propriétaire doit recevoir le préavis dans un délai de deux mois à dater de la\\ notification. Vous devez également fournir la preuve de la destination du bien et de votre lien de parenté avec le propriétaire. Vous devez également\\ fournir une estimation du coût de l'envoi du préavis et la date de réception du préavis. Vous devez également démontrer que le préavis a été reçu par le\\ propriétaire dans les délais prévus. Vous devez également fournir une estimation du coût de l'envoi du préavis et la date de réception du préavis. Vous\\ devez également démontrer que le préavis a été reçu par le propriétaire dans les délais prévus.\\
\graymidrule
\textbf{One-shot}: Vous pouvez envoyer votre demande de préavis au propriétaire en Wallonie par courriel, en vous adressant-le à son adresse email ou son\\ adresse postale. Vous pouvez également la faire parvenir par messagerie ou en colis recommé, s'il vous êtes dans un pays de l'Union européenne.\\Vous devez mentionner votre nom, votre adresse et votre nom de famille, ainsi que le nom du bien et le montant du loyer. Vous pouvez également\\ mentionner la durée de la demande et la raison de la demande. Vous pouvez également demander une réduction de loyer, si cela est le cas. Vous\\ pouvez également mentionner la date de paiement du loyer et le montant du loyer. Vous pouvez également mentionner si vous êtes en arrières de\\ paiement. Vous pouvez également mentionner si vous êtes en arrières de paiement et si vous êtes en situation de faillite. Vous pouvez également\\ mentionner si vous êtes en situation de faillite et si vous êtes en situation de faillite.\\
\graymidrule
\textbf{Few-shot}: Vous pouvez envoyer votre demande de préavis au propriétaire en Wallonie par courrier recommandé à la poste. Vous devez indiquer votre\\ nom, adresse et le nom du bien immobilier, ainsi que le montant du précompte immobilier et la durée de la demande. Vous devez également indiquer\\ si vous êtes en mesure de payer le précompte immobilier et si vous êtes en mesure de le payer. Vous pouvez également demander une réduction du\\ précompte immobilier. Vous devez fournir des preuves de votre situation familiale et de vos droits sur le bien immobilier. Vous pouvez également\\ demander une réduction du précompte immobilier si le bailleur ne prouve pas que le loyer a été fixé conformément aux dispositions du bail. Vous\\ pouvez également demander une réduction du précompte immobilier si le bailleur ne prouve pas que le bien est occupé par vous-même ou par vos\\ descendants, vos ascendants, collatéraux ou conjoint. Vous pouvez également demander une réduction du précompte immobilier si le bailleur ne\\ prouve pas que le bien est occupé par des collatéraux du troisième degré. Vous pouvez également demander une réduction du précompte immobilier\\ si le bailleur ne prouve pas que les travaux ont lieu dans les conditions et délais prévus.\\
\graymidrule
\textbf{Finetuned}: 1. Vous pouvez envoyer votre préavis à votre propriétaire par recommandé. Vous devez signer votre préavis et indiquer la date de réception.\\ Vous devez déposer votre préavis au bureau de paie du propriétaire. Le propriétaire doit recevoir votre préavis dans les 10 jours de sa réception. Il peut\\ aussi recevoir votre préavis par remise entre les mains du propriétaire et de l'autre. Le propriétaire doit signer votre préavis. Il doit aussi indiquer la date\\ de réception. Il doit vous envoyer votre préavis par recommandé si : vous envoyez votre préavis à votre propriétaire avant le début du mois qui suit le mois\\ dans lequel vous l'avez envoyé ; ou vous envoyez votre préavis à votre propriétaire après le début du mois qui suit le mois dans lequel vous l'avez envoyé ;\\ ou vous envoyez votre préavis à votre propriétaire pour y mettre fin à un moment précis. 2. Si vous ne respectez pas les délais de préavis, votre bail est\\ prolongé. Pour plus d'infos, voyez notre rubrique ``Fin et rupture du bail de 9 ans''.\\
\bottomrule
\end{tabular}
}
\caption{Example of long-form answers from WizardLM in different settings.}
\label{tab:qualitative_results}
\end{table*}

\section{Qualitative Results}
\Cref{tab:qualitative_results} showcases an example of the long-form answers that WizardLM-1.0 \citep{xu2023wizardlm} generates in response to a question sampled from the LLeQA test set, under its different configurations.

\begin{table}[t]
\centering
\resizebox{\columnwidth}{!}{%
\begin{tabular}{ll|r|ccc} 
\toprule
& \textbf{System} & \textbf{Cost/10K} & \textbf{Precision} & \textbf{Recall} & \textbf{F1} \\ 
\midrule
\multicolumn{6}{l}{\textbf{Baselines}} \\
& \textsc{first}          & n/a.          & 27.9 & 26.9 & 27.2 \\
& \textsc{random}         & n/a.          & 15.2 & 16.2 & 15.3 \\
\graymidrule
\multicolumn{6}{l}{\textbf{Proprietary models}} \\
& command-light           & unk.          & 19.2            &\underline{66.2} & 26.7 \\
& claude-instant-1.1      & \euro{} 45.8  & 37.7            & 42.6            & 38.1 \\
& gpt-3.5-turbo-16k-0613  & \euro{} 82.9  &\underline{47.1} & 50.8            & \underline{47.5} \\
& jurassic-2-mid          & \euro{} 252.4 & 32.2            & 56.6            & 38.6 \\
& claude-2.0              & \euro{} 326.5 & 34.8            & 45.7            & 37.9 \\
& command                 & \euro{} 337.6 & 25.4            & 24.0            & 24.5 \\
& jurassic-2-ultra        & \euro{} 378.9 & 35.0            & 62.5            & 41.1 \\
& gpt-4-0613              & \euro{} 757.2 & \textbf{62.5}   & \textbf{72.4}  & \textbf{65.5} \\

\graymidrule
\multicolumn{6}{l}{\textbf{Open-source models}} \\
& vicuna-33b-1.3          & n/a. & 25.0 & 42.5 & 30.5 \\
& wizardlm-30b-1.0       & n/a. & 26.5 & 53.6 & 33.3 \\
& guanaco-33b             & n/a. & 21.0 & 30.8 & 24.0 \\
& tulu-30b                & n/a. & 35.8 & 40.3 & 35.7  \\
\bottomrule
\end{tabular}
}
\caption{Zero-shot results for rationale extraction on LLeQA test set. We report the average cost of annotation per 10K samples for proprietary models accessible through an API.}
\label{tab:annotation_results}
\end{table}



\end{document}